# Singapore Soundscape Site Selection Survey ($S^5$): Identification of Characteristic Soundscapes of Singapore via Weighted *k*-means Clustering

**Kenneth Ooi [1,*], Bhan Lam [1], Joo Young Hong [2], Karn N. Watcharasupat [1], Zhen-Ting Ong [1] and Woon-Seng Gan [1]**

[1] School of Electrical and Electronic Engineering, Nanyang Technological University; wooi002@e.ntu.edu.sg (K. Ooi); bhanlam@ntu.edu.sg (B. Lam); karn001@e.ntu.edu.sg (K. N. Watcharasupat); ztong@ntu.edu.sg (Z.-T. Ong); ewsgan@ntu.edu.sg (W.-S. Gan)

[2] Department of Architectural Engineering, Chungnam National University; jyhong@cnu.ac.kr (J.Y. Hong)

* Correspondence: wooi002@e.ntu.edu.sg

**Abstract:** The ecological validity of soundscape studies usually rests on a choice of soundscapes that are representative of the perceptual space under investigation. For example, a soundscape pleasantness study might investigate locations with soundscapes ranging from "pleasant" to "annoying". The choice of soundscapes is typically researcher-led, but a participant-led process can reduce selection bias and improve result reliability. Hence, we propose a robust participant-led method to pinpoint characteristic soundscapes possessing arbitrary perceptual attributes. We validate our method by identifying Singaporean soundscapes spanning the perceptual quadrants generated from the "Pleasantness" and "Eventfulness" axes of the ISO 12913-2 circumplex model of soundscape perception, as perceived by local experts. From memory and experience, 67 participants first selected locations corresponding to each perceptual quadrant in each major planning region of Singapore. We then performed weighted *k*-means clustering on the selected locations, with weights for each location derived from previous frequencies and durations spent in each location by each participant. Weights hence acted as proxies for participant confidence. In total, 62 locations were thereby identified as suitable locations with characteristic soundscapes for further research utilizing the ISO 12913-2 perceptual quadrants. Audio-visual recordings and acoustic characterization of the soundscapes will be made in a future study.

## 1. Introduction

### 1.1. Background and Motivation

The idea of the soundscape approach, as defined in Part 1 of the International Standard for Soundscapes, ISO 12913-1:2014 [1], marks a paradigm shift from traditional noise control approaches to perception-driven approaches for planning and designing of sustainable urban acoustic environments. Considering soundscapes as "acoustic environments as perceived or experienced and/or understood by a person or people, in context" [1], soundscape studies have typically focused on the ecological validity and generalizability of their results by studying a range of soundscapes representative of the space of perceptual attributes under study [2–4].

However, for an arbitrary perceptual attribute, there may not necessarily exist standard examples of soundscapes possessing that attribute. Part 3 of the International Standard for Soundscapes, ISO 12913-3:2019 [5], defines a circumplex model with two orthogonal axes corresponding to the perception of "Pleasantness" and "Eventfulness" of a given soundscape, but does not provide examples of soundscapes that are quintessentially "pleasant" or "eventful". This is likely because of the need to perceive soundscapes *in context* — the perception of a given soundscape may differ from country to country or setting to setting. As a result, the identification of such soundscapes has often been based on external expert judgements made by soundscape researchers, or by a researcher-led choice of sets of environments with a sufficient variety of pre-defined objective characteristics under the assumption that they are representative of the corresponding perceptual attributes under study. A brief overview of soundscape studies employing either of these methods is shown in Table 1.

**Table 1.** Brief overview of soundscape studies employing researcher-driven choice of study areas.

| Study (Year) | Area(s) stimuli originated | Rationale for choice | Perceptual attribute(s) under study |
|---|---|---|---|
| Axelsson et al. (2010) [6] | London (United Kingdom), Stockholm (Sweden) | Variety in overall sound pressure level, types of sound sources | Agreement with 116 different affective attributes (for example, "pleasant" and "calm") |
| Axelsson (2015) [7] | Sheffield, London, Brighton (United Kingdom) | Variety in types of urban and peri-urban areas | Agreement with adjectives "pleasant", "vibrant", "eventful", "chaotic", "annoying", "monotonous", "uneventful", "calm" |
| Puyana Romero et al. (2016) [8] | Naples (Italy) | Variety in condition of road traffic flow | Perceived soundscape quality |
| Aumond et al. (2017) [9] | Paris (France) | Variety in types of urban areas | Pleasantness |
| Fan et al. (2017) [10] | Mixed (from Freesound [11]) | Variety in types of sound sources | Valence, arousal |
| Puyana Romero et al. (2019) [12] | Naples (Italy) | Variety in type of urban spaces | Agreement with adjectives "pleasant", "unpleasant", "monotonous", "exciting", "eventful", "uneventful", "chaotic", "calm" |
| Masullo et al. (2021) [13] | Mixed (from IADS-E database [14]) | Variety in types of urban sound sources | 2 sets of attributes (17 and 12 attributes) related to emotional salience |
| Hasegawa and Lau (2022) [15] | Singapore (Singapore) | Presence of common noise sources and greenery, resident demographic similarity | Pleasantness, eventfulness, satisfaction |

However, a choice of study areas or stimuli, if performed by the same researchers conducting the study, is at risk of selection and confirmation bias. Preventing investigator bias is crucial in building predictive models generalizable to multiple perceptual components, and for validation studies (such as the Soundscape Attributes Translation Project [16]) where a representative set of common stimuli is investigated under different contexts. This applies even if the external expert-guided decisions indeed span the perceptual space under study in post-hoc analysis. This argument is supported by findings in landscape assessments, where researcher-led decisions on landscape quality based on aesthetic features have been found to be less reliable than perception-based approaches with a group of human observers in participatory studies [17]. Since soundscape assessments share numerous parallels with landscape assessments [18], the potential lack of reliability conceivably applies to soundscape assessments as well.

Hence, an identification of characteristic soundscapes in local contexts by local experts, each of whom have experienced those soundscapes before, would arguably be more appropriate compared to that performed by soundscape researchers as external experts (possibly) unfamiliar with the local context [19]. Moreover, a replicable method to summarize the opinions of a sizeable population of local experts as participants in a participant-led process, *independent* of external expert-guided decisions, is desirable to provide sufficient blinding in the choice of study areas by soundscape researchers. Therefore, the overarching aim of this study is to execute such a method, by

- crowdsourcing opinions from a large sample of local experts via the administration of a standardized questionnaire,
- accounting for the reliability of each local expert in the sample via the numerical weighting of each opinion, and
- summarizing the crowdsourced opinions via an automatic, replicable clustering algorithm,

using regions in Singapore as a case study. In so doing, we hypothesize that our method will identify locations of characteristic soundscapes in Singapore possessing perceptual attributes of interest to ISO 12913-3:2019, even *without* using researcher opinions on what locations may possess such perceptual attributes.

### 1.2. Organization and Scope

This paper is organized as follows:

- **Section 2** provides a brief overview of work related to our study.
- **Section 3** provides a description of study area, the questionnaire used to elucidate locations from the participants of the study, and details on the weighted *k*-means clustering method we used to obtain locations of the characteristic soundscapes from the locations elucidated from the participants.
- **Section 4** presents the results of our proposed clustering method.
- **Section 5** analyzes the clusters and characteristic soundscapes obtained to validate the method.
- **Section 6** concludes our study and suggests possible directions for future work.

The scope of this work concerns only the identification of soundscapes possessing arbitrary perceptual attributes and the statistical validation of such an identification method. Empirical validation of the identified sites via in-situ observations and recordings are not in the scope of the present study.

## 2. Related Work

Apart from the studies highlighted in Table 1, an iconic study focusing on the systematic identification of characteristic soundscapes with local experts is the Urban Soundscapes of the World (USotW) study [20], which is part of a larger project that aims to identify and record soundscapes that are "full of life and exciting", "chaotic and restless", "calm and tranquil", and "lifeless and boring" in various cities around the world. These are descriptors located in each of the four quadrants generated by the "Pleasantness" and "Eventfulness" axes in the ISO 12913-3:2019 circumplex model, as observed in the original principal components analysis performed by Axelsson et al. [6] on 116 soundscape descriptors. A visual representation of the descriptors with respect to the circumplex model is shown in Figure 1.

Furthermore, Mediastika et al. [21] identified favorite public places in Indonesia with local experts who each identified three of their favorite locations, and then aggregated results by count and ranked them in descending order to identify the top locations as representative public places with unique sounds. In addition, Jeon et al. [22] proposed a participatory soundwalk approach with a group of acousticians and architects to identify positive and negative soundscapes along a designated soundwalk route, and grouped participants' chosen locations into 16 positions of interest, each with at least 5 chosen locations within a 10-meter radius.

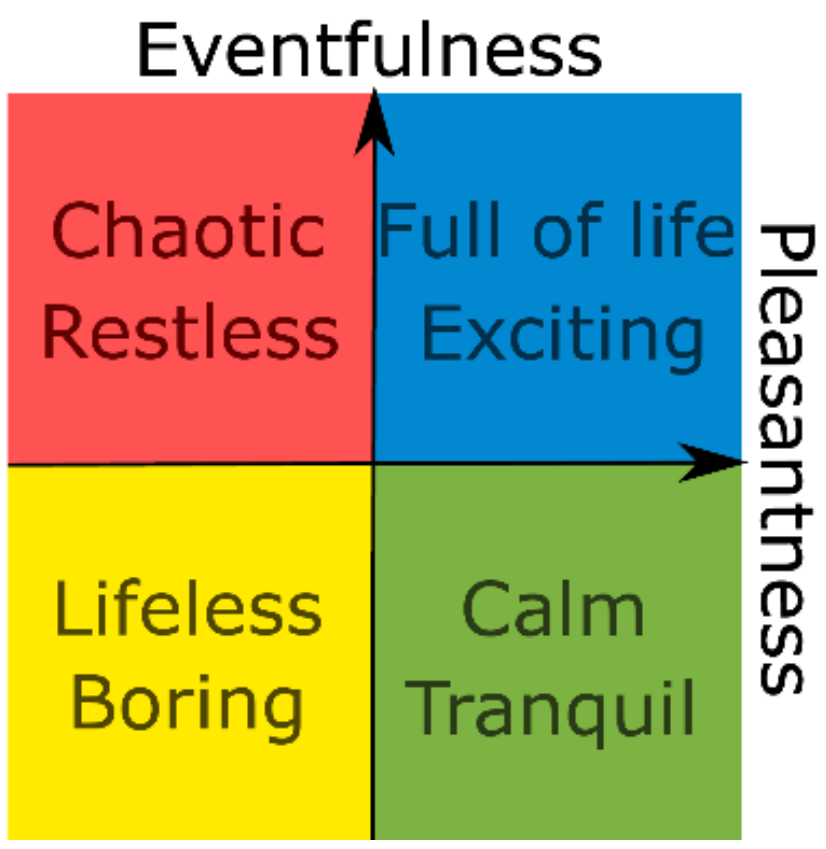

**Figure 1.** ISO 12913-3:2019 circumplex model of soundscape perception with descriptors for each quadrant drawn from [6].

However, these studies did not account for possibly varying levels or expertise and reliability in the participants surveyed. For instance, the original study conducted for the USotW project [20] performed clustering using the Google Maps Application Programming Interface (API) with equal weights on all points to obtain the locations with characteristic soundscapes, and Mediastika et al. [21] assigned exactly one vote for each location identified by a local expert. Local experts may differ in their levels of expertise, so accounting for this in the identification process is necessary to ensure greater reliability in the characteristic soundscapes subsequently identified. Hence, we propose a modification of weights to account for participant reliability in Section 3.4.

Density-based clustering methods have also been used to identify tourist landmarks. For example, Pla-Sacristán et al. [23] proposed the combination of a method they called "K-DBSCAN" to first identify places of interest, and another method they called "V-DBSCAN" to perform clustering of points into those places of interest. Their method used Global Positioning System (GPS) metadata of pictures (mostly taken by tourists) uploaded to a public website, but characteristic soundscapes might not necessarily be similar in nature to tourist landmarks depending on the perceptual attributes under study. For example, tourist landmarks might not necessarily have soundscapes that are "monotonous", even though that is a perceptual attribute of interest in ISO 12913-2:2018 [24]. A more general-purpose clustering method may thus be necessary (and suffice) for the purpose of identifying characteristic soundscapes, so we propose a modified *k*-means clustering algorithm in Section 3.5.

## 3. Materials and Methods

### *3.1. Study Area and Context*

For compatibility with the USotW database, we consider a similar collection of four sets of perceptual attributes for which we aim to identify characteristic soundscapes for. These four sets of attributes are namely "full of life and exciting", "chaotic and restless", "calm and tranquil", and "lifeless and boring". These attributes will be for soundscapes as perceived by local experts in the Singaporean context.

There is no widely-accepted definition of "local expert" in the context of soundscape research, so in line with the general idea that a "local expert" needs to be familiar with and living in the area under study [19], for the purposes of this study, we define a "local expert" to be a person who:

- has resided in Singapore for at least 10 years, or
- is a Singapore Tourism Board (STB) licensed tourist guide[1].

[1] STB licensed tourist guides are required to undergo the training described at https://www.stb.gov.sg/content/stb/en/assistance-and-licensing/licensing-overview/tourist-guide-licence.html (Last accessed: 11 May 2022) before obtaining their license.

Singapore, with a land area of 728 km², is also much larger than the city centers that were investigated for the USotW project. The questionnaire must thus be adapted to reduce selection bias in the characteristic soundscapes that will later be identified by the clustering method. To do so, we divide Singapore into the six planning regions as defined by the Urban Redevelopment Authority (URA)[2] of Singapore, administer a separate questionnaire for each region, and finally aggregate the points to perform the final clustering in Section 3.5. The planning regions, together with the names of some representative neighborhoods, are shown in Figure 2. The planning region officially designated as the "Central Area" in Figure 2 is also colloquially known as the "Central Business District" (CBD) or "CBD Area", and will henceforth be referred to as such to prevent confusion with the similarly-named "Central Region".

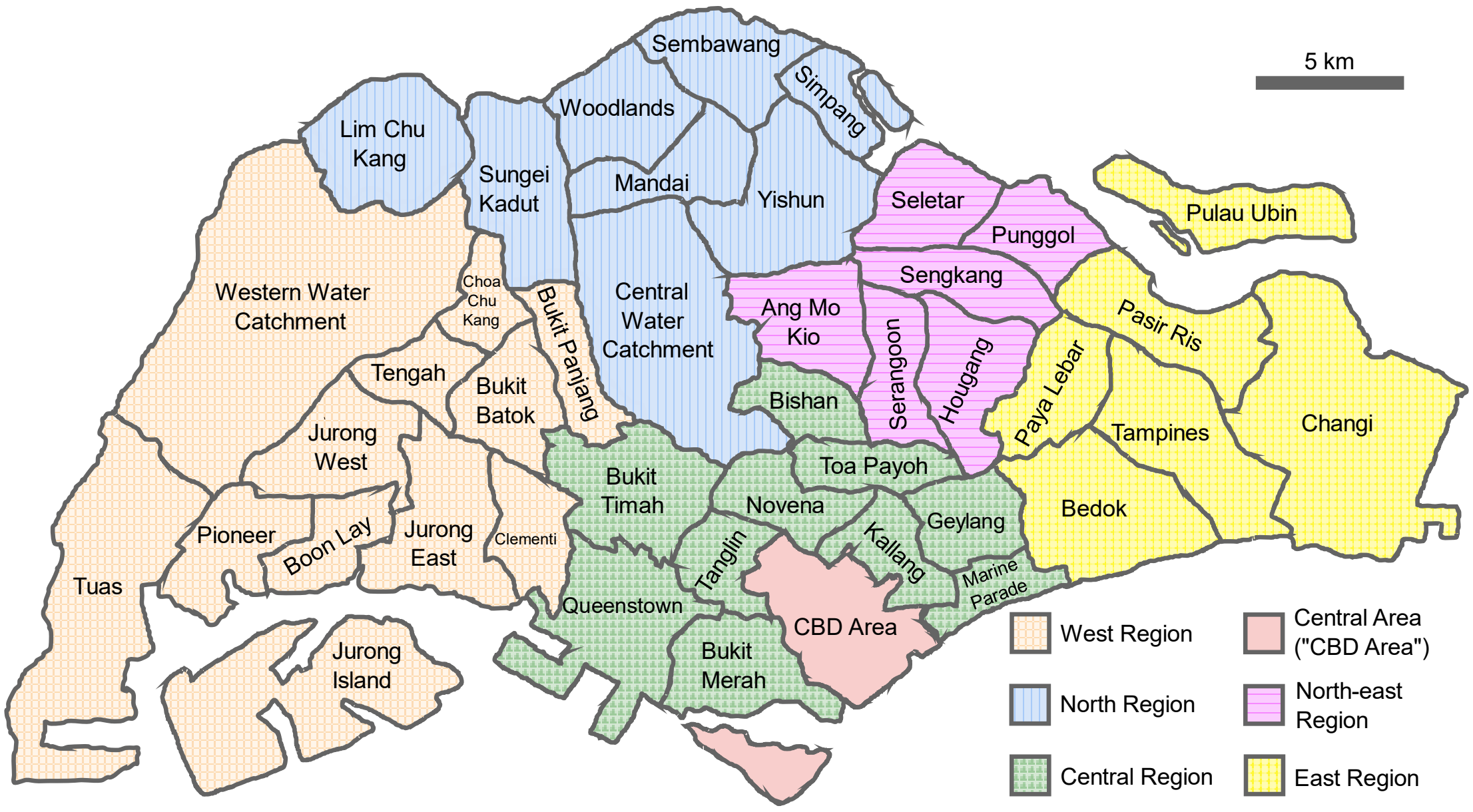

**Figure 2.** Major planning regions of Singapore as defined by the Urban Redevelopment Authority (URA) of Singapore (adapted from [25]).

### 3.2. Participants

Participants were recruited via email from a publicly available list of tourist guides licensed by the Singapore Tourism Board (STB), and via online messaging applications. All participants were fluent in English, and self-reported that they had no history of hearing loss, nor did they suspect that they had hearing loss. In total, 67 participants signed up for the study and were remunerated for their time.

After each participant signed up, we asked them if they were an STB-licensed tourist guide and obtained basic demographic information (age, gender, length of residence in Singapore, region of main residence) from them. We then administered the 21-item Weinstein Noise Sensitivity Scale (WNSS-21) [26], a standardized questionnaire to measure noise sensitivity in individuals. WNSS-21 was scored on a series of 5-point scales, and the total score for each participant was divided by 21 to obtain their normalized WNSS-21 score. The normalized WNSS-21 score ranges from 1 to 5, with 1 indicating low noise sensitivity and 5 indicating high noise sensitivity overall. A summary of the information obtained from all participants is shown in Figure 3.

[2] The Urban Redevelopment Authority (URA) of Singapore is the government agency in charge of land use planning and conservation in Singapore. Its official website is https://www.ura.gov.sg/ (Last accessed: 11 May 2022).

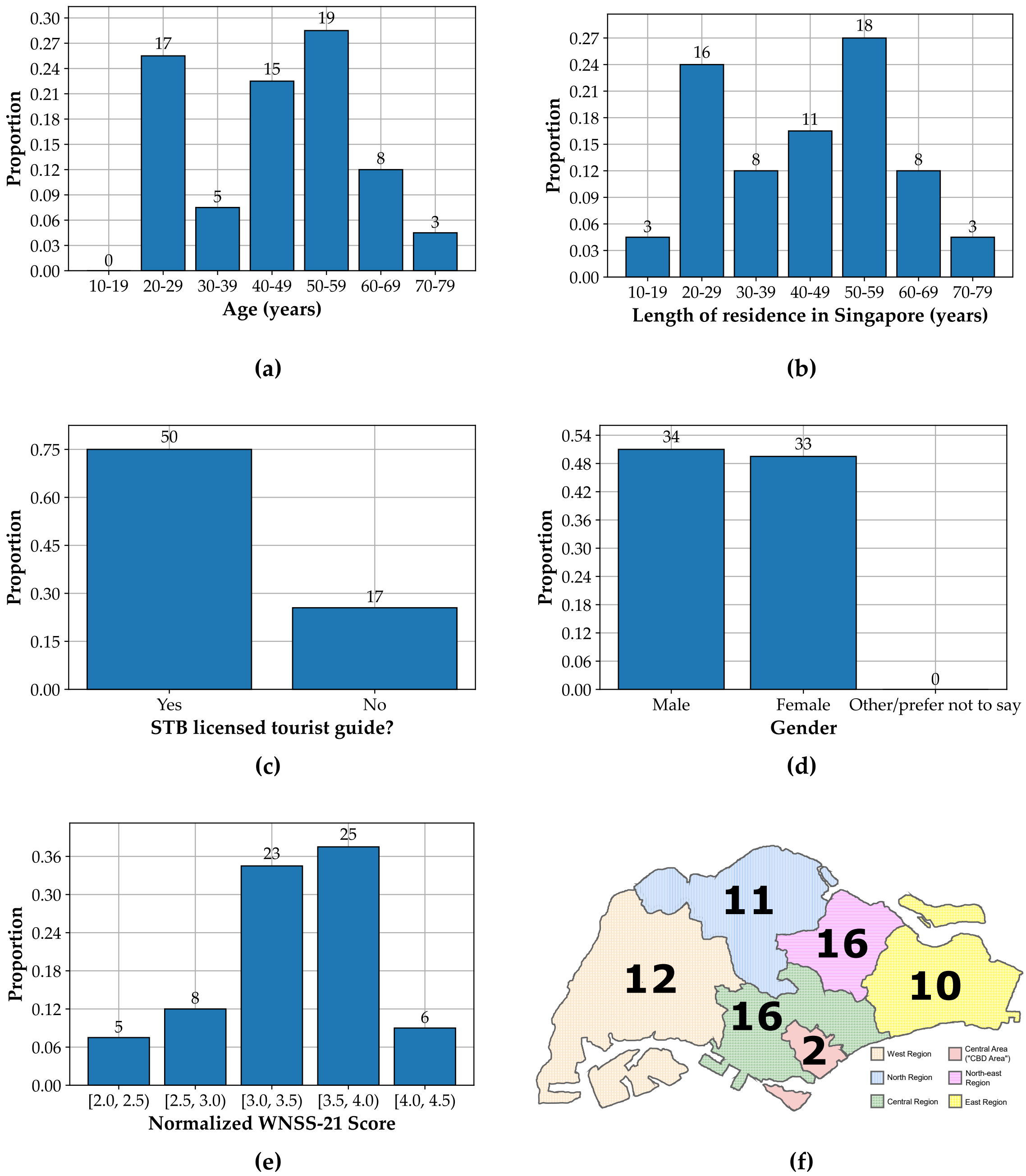

**Figure 3.** Aggregated demographic information of participants ($n$ = 67) in S[5], with numbers above bars in histograms denoting the exact number of participants in that bin: **(a)** Age distribution (Mean: 45.3, Standard Deviation (SD): 14.8); **(b)** Distribution of length of residence in Singapore (Mean: 43.9, SD: 15.7); **(c)** Whether participant is an STB licensed tourist guide; **(d)** Gender distribution; **(e)** Distribution of normalized WNSS-21 score on a scale of 1 to 5 (Mean: 2.62, SD: 0.50); **(f)** Distribution of participants' main residence by URA planning region.

### 3.3. Questionnaire

For each set of perceptual attributes in the four quadrants of Figure 1 ("full of life and exciting", "chaotic and restless", "calm and tranquil", "lifeless and boring"), each participant was asked to identify one location in each planning region of Singapore ("CBD Area", "Central Region", "East Region", "North Region", "North-east Region", "West Region" as shown in Figure 2) that they thought had a soundscape that best represented that set of perceptual attributes. This gave a total of 67 × 6 = 402 locations in Singapore for each set of perceptual attributes, and 402 × 4 = 1608 responses across all four sets of perceptual attributes. Participants were instructed to identify locations based on prior memory and experience, and only to identify locations which they had visited at least once before. Since the study was conducted amidst measures taken to minimize the spread of COVID-19, we additionally instructed participants to identify locations based on memory and experience *prior* to the implementation of these measures. Since COVID-19 mitigation measures have altered the perception of in-situ urban soundscapes [27], the instruction to identify locations *prior* to the implementation of these measures would make the locations applicable even *after* they have been lifted.

For each location identified, each participant was also asked to state how many times they had previously visited the area (or passed by it on foot). Their response to this question was then coded as a "frequency weight" according to Table 2.

**Table 2.** Coding table from number of times visited to frequency weight.

| Response (number of times visited) | Frequency weight |
|---|---|
| 1 to 3 | 1 |
| 4 to 6 | 2 |
| 7 to 9 | 3 |
| 10 or more | 4 |

Lastly, for each location identified, each participant was also asked to state the average duration per visit in minutes. The questionnaire was conducted via an online platform, and further details about its implementation can be found in Appendix A.

*3.4. Weight Assignment Accounting for Reliability*

As described in Section 3.3, each identified location was coupled with the frequency weight and average duration of each visit. Hence, the $i$-th response to the questionnaire can be characterized by a tuple $(x_i, f_i, t_i)$, where $x_i = (\varphi_i, \theta_i)$ denotes the coordinates of the location identified by the participant with $\varphi_i$ being the latitude and $\theta_i$ being the longitude in radians, $f_i$ being the frequency weight for $x_i$ as coded by Table 2, and $t_i$ being the average duration of each visit in minutes.

Since transient sound events significantly affect soundscape perception [28,29], and the probability of only experiencing transient events at a location increases with a decrease in time spent at the location, we consider locations where participants spent less total time as less reliable for the purposes of the *k*-means clustering. The lack of reliability in the location can also be interpreted as a lack of confidence on the part of the participant when identifying it. To quantify this reliability, we note that the product of the frequency weight and average duration per visit, $f_i t_i$, is a proxy measure of the overall time spent at each chosen location, and the reliability of a chosen location is correlated to the probability of the chosen location truly possessing the desired perceptual attributes. To transform the frequency-duration product $f_i t_i$ to a probability value between 0 and 1, we use the sigmoid function $\sigma(u) = \frac{1}{1+e^{-u}}$. Hence, the reliability measure of the coordinates $x_i$ of the chosen location is represented by assigning $x_i$ a weight $w_i$, where

$$w_i = \sigma(f_i t_i). \tag{1}$$

Here,

- $w_i$ denotes the reliability measure of the coordinates $x_i$ of the chosen location,
- $\sigma(\cdot)$ denotes the sigmoid function,
- $f_i$ denotes the frequency weight for $x_i$ as coded by Table 2, and
- $t_i$ denotes the average duration of each visit to $x_i$ in minutes.

As an empirical validation of the reliability of this method, we plot the weights $w_i$ for all locations identified by the participants for each set of perceptual attributes in ascending order in Figure 4. Since $f_i t_i > 0$, the minimum weight for any point is 0.5, and we can see that the weights cover almost the entire range of possible values between 0.5 and 1. In addition, the weights for locations identified as "lifeless and boring" tended to be lower than that for the other three sets of perceptual attributes, as can be seen from the reduced area under the "lifeless and boring" curve of 0.70 in Figure 4 as compared to 0.91, 0.81, and 0.89 for the areas under the "full of life and exciting", "chaotic and restless", and "calm and tranquil" curves, respectively. This indicates that participants were less confident with their choice of "lifeless and boring" locations, which agrees with an observation made in the USotW study [20], whose participants generally found "lifeless and boring" locations hardest to select.

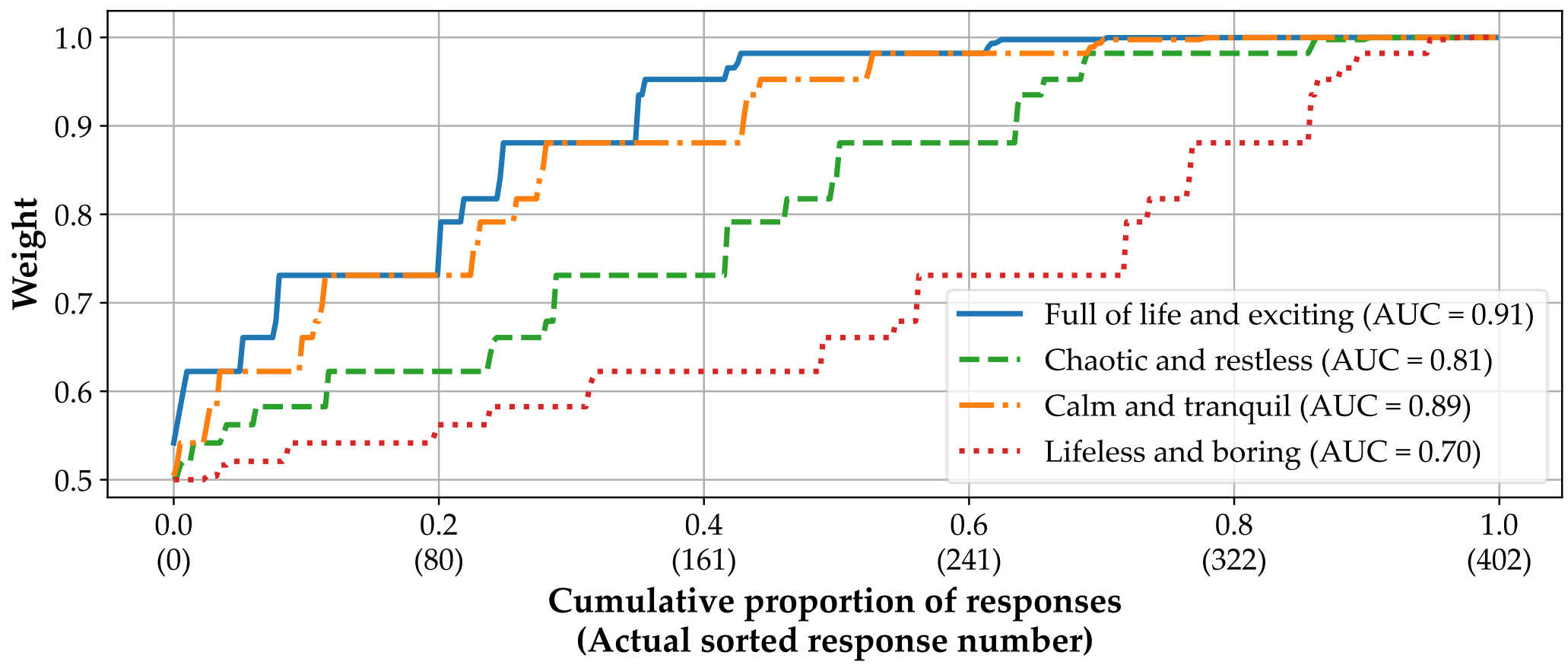

**Figure 4.** Weights used for clustering against cumulative proportion of responses (and corresponding actual sorted response numbers in increasing order) at locations identified for each set of perceptual attributes (labelled with area under curve (AUC)).

### *3.5. Clustering Method*

After assigning the weights $w_i$ computed via Equation (1) to all locations $x_i = (\varphi_i, \theta_i)$ represented as latitude-longitude pairs, a weighted *k*-means clustering was performed on each of the 402 locations identified by the participants in Singapore for each set of perceptual attributes. Consequently, the dataset used for clustering consisted of the following input variables:

- the latitude of the location $\varphi_i$,
- the longitude of the location $\theta_i$, and
- the weight $w_i$ associated with the location.

The clustering method was identical for each set of perceptual attributes, and followed the standard *k*-means clustering method (outlined in Section 3.5.1) with modifications described in Sections 3.5.2 to 3.5.4.

#### 3.5.1. Standard *k*-means Clustering Method

The standard *k*-means clustering method is an unsupervised learning technique typically used to group unlabeled data points into clusters such that the within-cluster variance is minimized [30]. This simultaneously minimizes the overall distance between points and their closest cluster center (computed as the mean of all points in the cluster), such that cluster centers effectively represent their clusters. The algorithm used to perform the standard *k*-means clustering method iteratively updates the cluster centers and points in each cluster in an alternating manner, and one possible implementation of it is given as Algorithm 1. As an example of its application in soundscape analysis, Flowers et al. [31] used the standard *k*-means clustering algorithm to cluster soundscape recordings based on eight acoustic indicators and analyzed the clusters to reveal spatiotemporal correlations within the clusters.

**Algorithm 1.** Standard $k$-means clustering method

**Inputs:** Set of $n$ points $X = \{x_1, x_2, \dots, x_n\}$ to be clustered
Number of clusters $k$

**Outputs:** Set of $k$ cluster centers $C = \{c_1, c_2 \dots, c_k\}$
Set of $k$ clusters $\mathcal{C} = \{C_1, C_2, \dots, C_k\}$, where $C_i \cap C_j = \emptyset$ for all $i \neq j$ and $\bigcup_{i=1}^{k} C_i = X$

**Initialization:**
$C \leftarrow$ Random $k$-element subset of $X$; // Each subset chosen with equal probability.

**while** not converged **do** // Convergence is reached when $\mathcal{C}$ remains unchanged for 1 iteration of the "**while**" loop.

  **for** $i = 1$ **to** $n$ **do**

    $C_i \leftarrow \left\{x_j : \underset{c \in C}{\operatorname{argmin}} \left(x_j - c\right)^2 = c_i\right\}$; // Assign points to the cluster whose center they are closest in Euclidean distance to.

    $c_i \leftarrow \frac{1}{|C_i|} \sum_{\{j : x_j \in C_i\}} x_j$ ; // Update cluster center as mean of all points in cluster.

**Return:** $C, \mathcal{C}$

3.5.2. Modification 1: Haversine Distance Metric

Instead of the Euclidean distance used in Algorithm 1, we use the haversine distance metric to account for the curvature of the Earth, since the haversine distance between any two points on a sphere is the shortest distance between them *when travelling on its surface*. Since the points chosen by the participants on the map were given in latitude-longitude pairs on the surface of the Earth, we assume a spherical Earth with radius $R = 6371$, in kilometers, and compute the haversine distance between any two points with latitude-longitude pairs $x_1 = (\varphi_1, \theta_1)$ and $x_2 = (\varphi_2, \theta_2)$ as

$$d(x_1, x_2) = 2R \tan^{-1}\left(\frac{\sqrt{a}}{\sqrt{1-a}}\right), \tag{2}$$

where

$$a = \sin^2\left(\frac{\varphi_2 - \varphi_1}{2}\right) + \cos(\varphi_1)\cos(\varphi_2)\sin^2\left(\frac{\theta_2 - \theta_1}{2}\right). \tag{3}$$

Here,

- $d(x_1, x_2)$ denotes the haversine distance between two points $x_1 = (\varphi_1, \theta_1)$ and $x_2 = (\varphi_2, \theta_2)$ on a sphere,
- $R$ denotes the sphere radius (with $R = 6371$ (in km) assuming a spherical Earth),
- $\varphi_1$ denotes the latitude of the point $x_1$ on the sphere,
- $\theta_1$ denotes the longitude of the point $x_1$ on the sphere,
- $\varphi_2$ denotes the latitude of the point $x_2$ on the sphere, and
- $\theta_2$ denotes the longitude of the point $x_2$ on the sphere.

By convention, we also assume $\varphi_1, \varphi_2 \in \left[-\frac{\pi}{2}, \frac{\pi}{2}\right]$ and $\theta_1, \theta_2 \in (-\pi, \pi]$.

3.5.3. Modification 2: Cluster Center Initialization with $k$-means++

Since the standard $k$-means algorithm can result in suboptimal solutions upon convergence, Arthur and Vassilvitskii [32] proposed a cluster initialization method known as "$k$-means++" that chooses the initial cluster centers one by one, with a point having a reduced probability of being chosen as a cluster center the nearer it is to existing cluster centers. The probability reduction is proportional to the squared distance of a point to its nearest cluster center. This is in contrast to the standard cluster center initialization method of picking $k$ points from a uniform distribution over all points. The initialization method is summarized in Algorithm 2.

**Algorithm 2.** Cluster center initialization with *k*-means++ (adapted from [32])

**Inputs:** Set of $n$ points $X = \{x_1, x_2, \dots, x_n\}$ to be clustered
Number of clusters $k$
Distance metric $d(\cdot,\cdot)$
**Output:** Set of $k$ initial cluster centers $C = \{c_1, c_2 \dots, c_k\}$
**Initialization:**
$C \leftarrow \emptyset$ ; // Initialize set of cluster centers as empty set.
$\mathbf{r} \leftarrow [\infty, \dots, \infty] \in \mathbb{R}^n$ ; // $\mathbf{r}[m]$ is the distance from the point $x_m \in X$ to its closest center in $C$.
$\mathbf{p} \leftarrow \left[\frac{1}{n}, \dots, \frac{1}{n}\right] \in \mathbb{R}^n$ ; // $\mathbf{p}[m]$ is the probability that the point $x_m \in X$ is chosen as an initial cluster center in $C$. Probabilities are initialized uniformly.
**for** $i = 1$ **to** $k$ **do**
$c_i \overset{\sim\mathbf{p}}{\leftarrow} X$ ; // Choose $c_i$ as a random point from $X$, where $x_m$ is chosen with probability $\mathbf{p}[m]$.
$C \leftarrow C \cup \{c_i\}$ ; // Append $c_i$ to the set of cluster centers.
**for** $j = \mathbf{1}$ **to** $n$ **do**
$\mathbf{r}[j] \leftarrow \min_{c \in C} d(c, x_j)$ ; // Update the distance from each point to its nearest cluster center in C.
**for** $j = 1$ **to** $n$ **do**
$\mathbf{p}[j] \leftarrow \frac{(\mathbf{r}[j])^2}{\sum_{l=1}^{n} (\mathbf{r}[l])^2}$ ; // Update the new probability for each point using **r**.
**Return:** $C$

Essentially, the *k*-means++ initialization method spaces out the initial cluster centers in an adaptive fashion across the dataset, thereby improving both convergence speed and the optimality of clusters obtained via *k*-means clustering. It has also been shown to achieve similar improvements in results for any distance metric [33], such as the haversine distance metric described in Section 3.5.2. Therefore, we replaced the initialization step of Algorithm 1 (where $C$ is assigned as a random *k*-element subset of $X$) with the output of Algorithm 2. Alternative methods to improve the convergence speed and optimality of clusters obtained via *k*-means clustering exist, such as Meta-Heuristics Tabu Search with Adaptive Search Memory (MHTSASM) [34], but require the tuning or setting of additional hyperparameters. Therefore, we have opted for *k*-means++ since it does not rely on any hyperparameters beyond the number of clusters *k* that is required by default.

#### 3.5.4. Modification 3: Cluster Center Computation with Weighted Means

To reflect the reliability of the locations identified by each participant, in each iteration of the *k*-means clustering algorithm, we computed each cluster center as the weighted mean of all points $x_i$ in that cluster, with the weights being the values $w_i$ computed via Equation (1). In other words, if we denote with $C_i$ the set of all points in the $i$-th cluster, then its cluster center $c_i$ is computed as

$$c_i = \frac{S_i}{W_i} = \frac{\text{Weighted sum}}{\text{Sum of weights}}, \tag{4}$$

where

$$S_i = \sum_{\{j : x_j \in C_i\}} w_j x_j, \tag{5}$$

and

$$W_i = \sum_{\{j: x_j \in C_i\}} w_j. \tag{6}$$

Here,

- $c_i$ denotes the cluster center of the $i$-th cluster,
- $C_i$ denotes the set of all points in the $i$-th cluster,
- $x_j$ denotes the point with index $j$,
- $w_j$ denotes the weight of the point $x_j$ computed via Equation (1), and
- $\{j: x_j \in C_i\}$ denotes the set of indices of all the points in the $i$-th cluster.

Notice that Equation (4) simplifies to the standard mean if all the weights are equal (in other words, if $w_j = \frac{1}{|C_i|}$ for all $j$ as in Algorithm 1). Using the weighted mean in the computation of cluster centers effectively moves the cluster centers towards points with higher weights, which are themselves proxies for reliability. The *k*-means clustering algorithm is still guaranteed to converge even with the use of the weighted mean, because each weight $w_i$ can also be interpreted as a ratio of multiple points located at $x_i$. For example, if two points $x_1$ and $x_2$ respectively have weights 0.5 and 0.75 in the weighted mean computation of cluster centers, it would be identical to having 2 points at $x_1$ and 3 points at $x_2$ in the standard mean computation of cluster centers.

## 4. Results

### *4.1. Optimal Number of Clusters*

For each set of perceptual attributes in the four quadrants of Figure 1, we ran the weighted *k*-means clustering algorithm described in Section 3.5 for 100 times, using different seeds each time, at values of $k$ ranging from 2 to 20 (inclusive) to determine an optimal number of clusters for each set of perceptual attributes. Since there were 402 points per set of perceptual attributes, we did not exceed $\sqrt{402} \approx 20$ clusters to ensure that, on average, the number of points per cluster is at least the same as the number of clusters.

To determine an optimal number of clusters, we used the Dunn index $\alpha(\mathcal{C}; k)$, which depends on the number of clusters $k$ and the set of clusters $\mathcal{C} = \{C_1, C_2, \dots, C_k\}$, where $C_i$ is the set of all points in cluster $i$. Specifically, the Dunn index for a distance metric $d(\cdot,\cdot)$ is defined by

$$\alpha(\mathcal{C}; k) = \frac{\min\limits_{1 \le i < j \le k} \left[ \min\limits_{x \in C_i, y \in C_j} d(x, y) \right]}{\max\limits_{1 \le i \le k} \left[ \max\limits_{x, y \in C_i} d(x, y) \right]} = \frac{\text{Minimum inter-cluster distance}}{\text{Maximum intra-cluster distance}}. \tag{7}$$

Here,

- $\mathcal{C} = \{C_1, C_2, \dots, C_k\}$ denotes the set of clusters,
- $k$ denotes the number of clusters,
- $C_i$ denotes the set of all points in the $i$-th cluster, and
- $d(x, y)$ denotes the denotes the distance between two points $x$ and $y$.

The Dunn index is a standard measure of cluster separation [35]. It has also been used in spatiotemporal cluster analysis of urban acoustic environments based on sound pressure level parameters in Barcelona, Spain [36] and for the clustering and classification of urban park soundscapes in Seoul, Korea based on acoustic indicators [37]. Furthermore, the Dunn index takes a value in the range $[0, \infty)$, and has a higher value when clusters are small (thus decreasing the maximum intra-cluster distance) and far apart (thus increasing the minimum inter-cluster distance). A higher Dunn index thus indicates a more optimal clustering, since small and separated clusters are usually desired from clustering methods.

The highest Dunn index obtained from the 100 runs for each value of $k \in \{2, 3, \dots, 20\}$ and each set of perceptual attributes is shown in Figure 5. From Figure 5, we can also see that the highest Dunn index is attained when there are 15 clusters for the attribute "full of life and exciting", 14 clusters for the attribute "chaotic and restless", 15 clusters for the attribute "calm and tranquil", and 18 clusters for the attribute "lifeless and boring". Therefore, these are the number of characteristic soundscapes to be extracted from the responses to the questionnaire in Section 3.3 for each set of perceptual attributes.

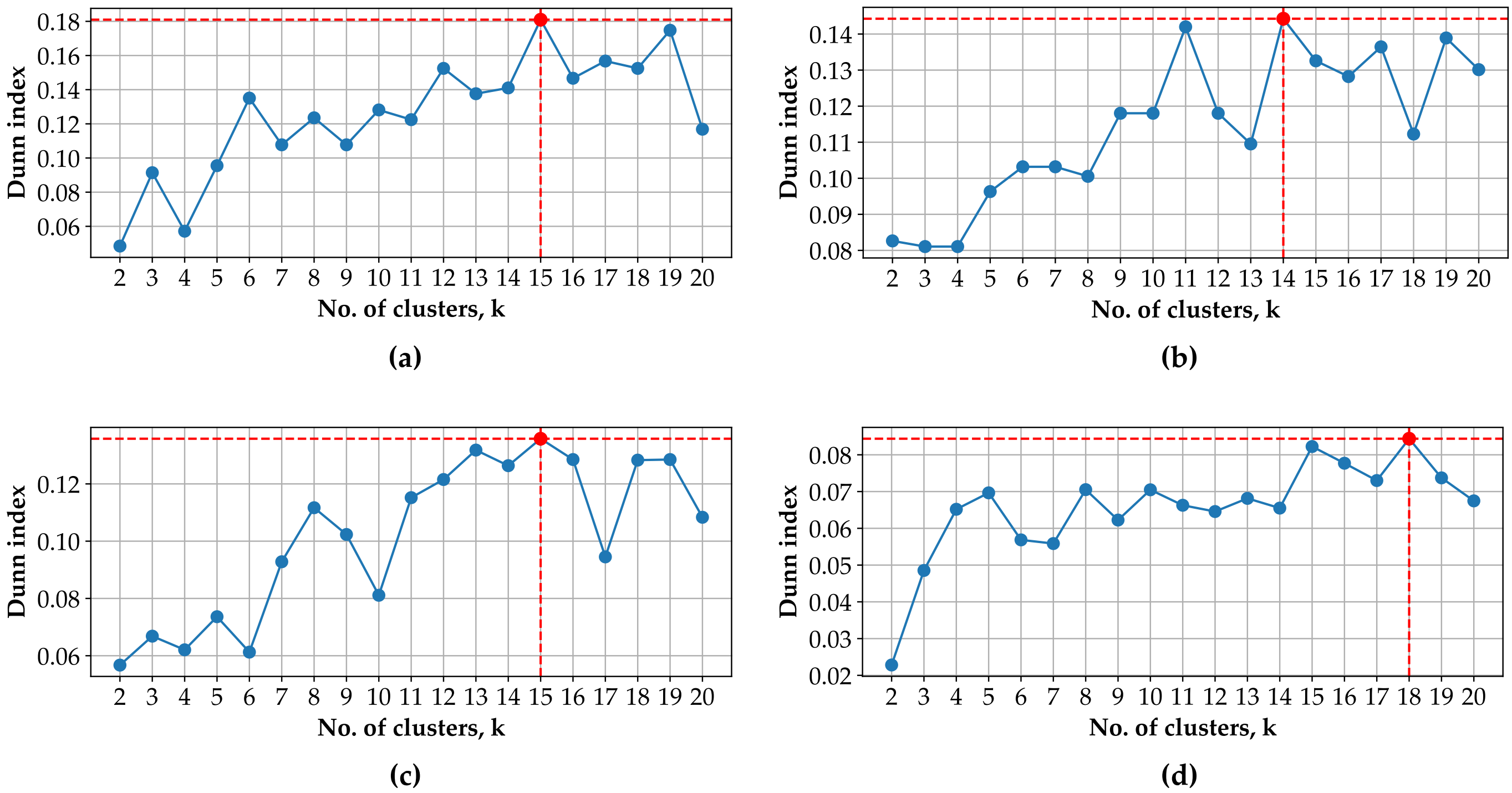

**Figure 5.** Maximum Dunn index values by number of clusters (with optimal number represented by red circles and dashed lines) for points labelled as: **(a)** full of life and exciting; **(b)** chaotic and restless; **(c)** calm and tranquil; **(d)** lifeless and boring.

### *4.2. Cluster Centers*

As mentioned in Section 4.1, we extracted the results of weighted *k*-means clustering with 15, 14, 15, and 18 clusters for each of the sets of perceptual attributes "full of life and exciting", "chaotic and restless", "calm and tranquil" and "lifeless and boring", respectively. The locations chosen by each participant and the resultant cluster centers obtained from the weighted *k*-means clustering algorithm described in Section 3.5 are shown in Figure 6, superimposed on a map of Singapore.

## 5. Discussion

### *5.1. Distribution of Cluster Centers*

Given the clusters shown in Figure 6, we can see that all six planning regions of Singapore each contain at least one cluster center, which indicates that the clusters have indeed covered the range of points chosen by the participants well. The distribution of cluster centers also shows that all six planning regions contain characteristic soundscapes possessing all four sets of perceptual attributes, albeit in unequal number.

From Figure 6, we can additionally observe that the number of cluster centers in the CBD Area is the least among the six regions (4 in total across all four sets of perceptual attributes). This is likely because it is the smallest planning region in Singapore, thus leading to many points being close together and being able to be represented well by a single cluster center. On the other hand, the number of cluster centers in the West Region is the greatest among the six regions (16 in total across all four sets of perceptual attributes), because the West Region is the largest planning region in Singapore and requires a greater number of cluster centers to represent.

The Central, East, and North Regions have 12, 10, and 14 cluster centers across all four sets of perceptual attributes, but of note is the North-east Region, which only has 6 cluster centers across the four sets of perceptual attributes despite it being roughly the same size as the Central and East Region. This is possibly due to several clusters having points that are contained in both the North-east and Central Regions, and the weighted mean computation caused the cluster centers to be located in the Central Region instead of the North-east Region. This is especially evident in Figures 6(a) and 6(b), which respectively have 1 and 2 such clusters whose cluster centers are in the Central Region, but are very close to the border of the Central and North-east Regions.

Lastly, we can see from Figure 6(d) that the points identified by the participants to be "lifeless and boring" are spread out most sparsely among the four sets of perceptual attributes, followed in order by the points identified to be "calm and tranquil", "chaotic and restless", and "full of life and exciting". This visual observation is supported by the fact that participants tended to have lower confidence in their identification of "lifeless and boring" locations in Figure 4, as well as the Dunn indices computed in Figure 5 for each of the four sets of perceptual attributes, where the highest Dunn indices for "lifeless and boring", "calm and tranquil", "chaotic and restless", and "full of life and exciting" are in increasing order from 0.084, 0.136, 0.144, and 0.181.

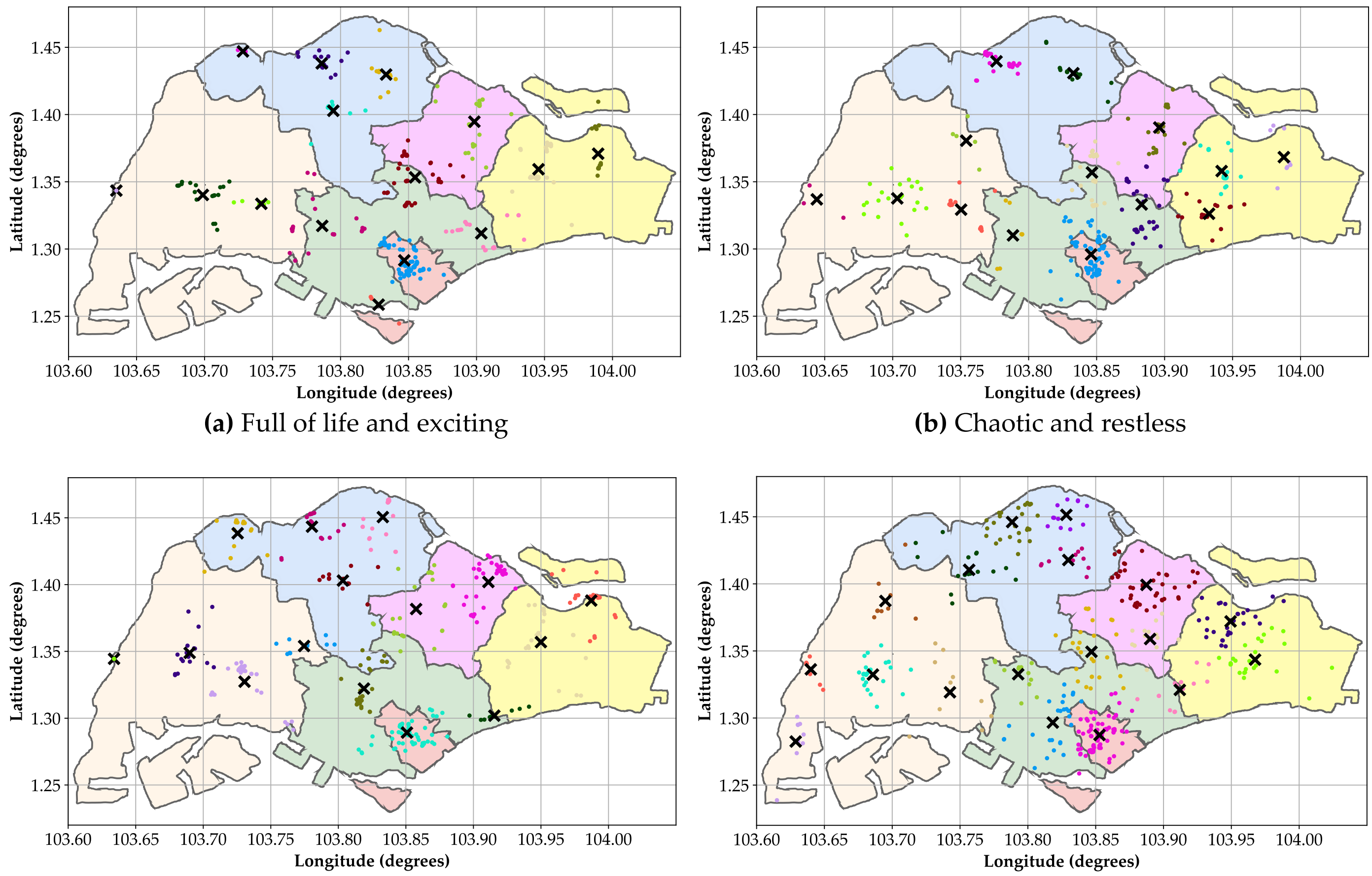

**(a)** Full of life and exciting

**(b)** Chaotic and restless

**(c)** Calm and tranquil

**(d)** Lifeless and boring

**Figure 6.** Cluster centers (black crosses) and points (marked with a unique color for each cluster), superimposed on a map of Singapore and representing locations considered by participants to be the most: **(a)** full of life and exciting; **(b)** chaotic and restless; **(c)** calm and tranquil; **(d)** lifeless and boring. Regions are color-coded in the same manner as Figure 2.

*5.2. Characteristic Soundscapes*

Given the cluster centers, it remains to choose locations with characteristic soundscapes in Singapore for each set of perceptual attributes. It may not be realistic to choose the locations of cluster centers as the exact locations of the characteristic soundscapes, because the cluster centers may be located sufficiently far away from the locations chosen by the participants in the Section 3.3 questionnaire to have soundscapes significantly different from the locations chosen by the participants. As an illustration, suppose a cluster contains exactly two points located at busy traffic intersections on opposite sides of a park, like that shown in Figure 7. If both points have exactly the same weights, the cluster center would be located in the middle of the park, which would most likely have different environmental and perceptual characteristics from busy traffic intersections.

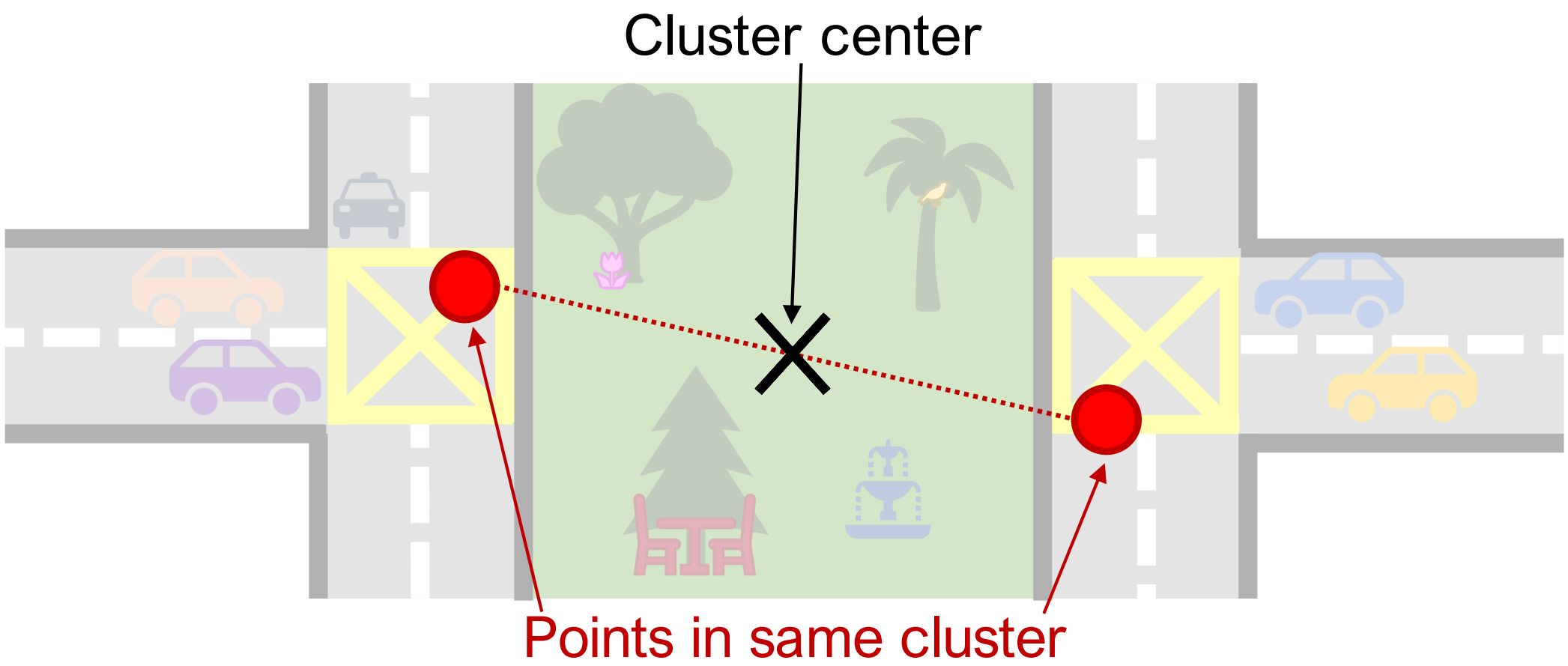

**Figure 7.** Illustration of how environmental and perceptual characteristics of cluster center locations may differ from actual locations identified by participants. Here, the actual locations identified are busy traffic intersections, but the cluster center is in the middle of a park.

In light of this concern, it is preferable to have actual points chosen by the participants as characteristic soundscapes to prevent an unexpected domain shift as a result of the identification procedure. Hence, we chose the point closest to each cluster center, in terms of the haversine distance, as the location with the characteristic soundscape for each cluster. No weights were applied for this distance computation, since the cluster centers already reflect the weights of the points in each cluster from the weighted *k*-means clustering algorithm. The resultant points chosen via this method are listed in Tables 3 to 6, with coordinates and a short textual description of the exact location.

We can see from Table 3 that the characteristic soundscapes considered "full of life and exciting" consist primarily of shopping malls, bus interchanges, and mass rapid transit (MRT) train stations. From Table 4, the characteristic soundscapes considered "chaotic and restless" contain comparatively more wide roads and intersections, whereas from Table 5, the characteristic soundscapes considered "calm and tranquil" are overwhelmingly in parks and nature reserves. Lastly, from Table 6, a number of characteristic soundscapes considered "lifeless and boring" are located along smaller roads, residential areas, and housing blocks. This shows that that there is a reasonably clear distinction in the locations whose soundscapes evoke the perceptual constructs "full of life and exciting", "chaotic and restless", "calm and tranquil", and "lifeless and boring" in the participants of this study, even without the use of any researcher input to the automatic, modified *k*-means algorithm used to identify these characteristic soundscapes. Hence, this observation backs our initial hypothesis in Section 1.1 for our proposed method that the locations have soundscapes possessing distinct perceptual attributes of interest.

**Table 3.** Details for characteristic soundscapes considered "full of life and exciting".

| ID | Region | Latitude (degrees) | Longitude (degrees) | Description |
|---|---|---|---|---|
| A01 | CBD | 1.291598203 | 103.8465300 | Opposite Clarke Quay Shopping Mall |
| A02 | East | 1.354207500 | 103.9435079 | Tampines Bus Interchange |
| A03 | East | 1.363875914 | 103.9914004 | Changi Airport Terminal 1 |
| A04 | Central | 1.263173177 | 103.8228356 | VivoCity Shopping Mall |
| A05 | Central | 1.301498905 | 103.9049564 | Parkway Parade Shopping Mall |
| A06 | Central | 1.311034361 | 103.7943141 | Holland Village Market & Food Centre |
| A07 | Central | 1.350677442 | 103.8494603 | Junction 8 Shopping Mall |
| A08 | North | 1.404012379 | 103.7934915 | Singapore Zoo (Ah Meng Restaurant) |
| A09 | North | 1.429740500 | 103.8351859 | Yishun MRT Station |
| A10 | North | 1.437221700 | 103.7861714 | Woodlands MRT Station |
| A11 | North | 1.446914441 | 103.7301914 | Sungei Buloh Wetland Reserve (Mangrove Boardwalk) |
| A12 | North-east | 1.392070753 | 103.8956615 | Compass One Shopping Mall |
| A13 | West | 1.333243872 | 103.7414451 | Jurong East MRT Station |
| A14 | West | 1.336767900 | 103.6941672 | Jurong West Sports Hall (facing Jurong West Street 93) |
| A15 | West | 1.343433486 | 103.6351438 | Raffles Marina |

**Table 4.** Details for characteristic soundscapes considered "chaotic and restless".

| ID | Region | Latitude (degrees) | Longitude (degrees) | Description |
|---|---|---|---|---|
| B01 | CBD | 1.300102657 | 103.8459222 | Handy Road (Opposite Plaza Singapura Shopping Mall) |
| B02 | East | 1.324737167 | 103.9306484 | Bedok Interchange Hawker Centre |
| B03 | East | 1.359156559 | 103.9407174 | Tampines Central 7 (Road) |
| B04 | East | 1.364476558 | 103.9915721 | Changi Airport Terminal 1 |
| B05 | Central | 1.310991457 | 103.7947432 | Holland Village Market & Food Centre |
| B06 | Central | 1.335196760 | 103.8844747 | Harrison Industrial Building |
| B07 | Central | 1.350930707 | 103.8480879 | Bishan MRT Station |
| B08 | North | 1.429664842 | 103.8341680 | S-11 Yishun 744 Hawker Centre |
| B09 | North | 1.442881682 | 103.7756387 | Opposite SPC Admiralty (Petrol Station) |
| B10 | North-east | 1.391455032 | 103.8955306 | Sengkang Bus Interchange |
| B11 | West | 1.333995645 | 103.6346393 | Intersection of Tuas West Drive & Pioneer Road |
| B12 | West | 1.337641500 | 103.7036367 | Intersection of Jurong West Street 63 & Jurong West Street 64 |
| B13 | West | 1.334852761 | 103.7461658 | IMM Shopping Mall |
| B14 | West | 1.379686712 | 103.7606068 | Intersection of Woodlands Road & Choa Chu Kang Road |

However, this distinction may not perfect, because there are coordinates which share almost identical locations, namely two corresponding to Changi Airport Terminal 1 (A03 and B04) and two corresponding to Holland Village Market & Food Centre (A06 and B05). Nonetheless, the exact coordinates in the tables are different, so it is also possible that the soundscapes at the exact coordinates differ from each other.

### *5.3. Limitations*

Even though we performed a spatial clustering of points to obtain the characteristic soundscapes, there is also a need to consider the time of day that the soundscape at that location exhibits the perceptual attributes specified in the questionnaire, since the sound source composition of each soundscape may be different at different times of day [38], and may thus affect the perception. Although there was a part of the questionnaire in Section 3.3 for participants to report reasons for their choice of location and characteristic *times* on top of the location of the characteristic soundscapes, not all participants did so. Hence, we were unable to carry out a spatiotemporal clustering approach (with time as an additional input to the weighted *k*-means clustering algorithm on top of the latitude-longitude pairs) in as rigorous a manner as the purely spatial approach (using only latitude-longitude pairs) described in Section 3.5.

Follow-up interviews could be conducted with the participants who chose the locations of the characteristic soundscapes to elucidate a range of suitable time periods matching the perceptual attributes under consideration. Alternatively, a separate study with other local experts could be conducted to identify these characteristic times, *given* the locations of the characteristic soundscapes.

**Table 5.** Details for characteristic soundscapes considered "calm and tranquil".

| ID | Region | Latitude (degrees) | Longitude (degrees) | Description |
|---|---|---|---|---|
| C01 | CBD | 1.290393318 | 103.8510017 | National Gallery Singapore (Museum) |
| C02 | East | 1.362116882 | 103.9467685 | Tampines Eco Green Park |
| C03 | East | 1.388507653 | 103.9884539 | Changi Beach Park |
| C04 | Central | 1.301187849 | 103.9156572 | East Coast Park (Area C) |
| C05 | Central | 1.320559055 | 103.8162867 | Botanic Gardens Eco Lake |
| C06 | North | 1.404355599 | 103.8036195 | Upper Seletar Reservoir |
| C07 | North | 1.441122864 | 103.7228199 | Sungei Buloh Wetland Reserve (Buloh Besar River) |
| C08 | North | 1.446485425 | 103.7805025 | Admiralty Park |
| C09 | North | 1.451390510 | 103.8405410 | Sembawang Park |
| C10 | North-east | 1.374836700 | 103.8455383 | Ang Mo Kio Town Garden West |
| C11 | North-east | 1.408367898 | 103.9072628 | Punggol Waterway Park |
| C12 | West | 1.334123398 | 103.7277980 | Jurong Lake Gardens |
| C13 | West | 1.344436500 | 103.6339522 | Johor Straits Lighthouse |
| C14 | West | 1.348941989 | 103.6876865 | NTU Sports and Recreation Centre |
| C15 | West | 1.354816300 | 103.7762985 | Bukit Timah Hill Summit |

**Table 6.** Details for characteristic soundscapes considered "lifeless and boring".

| ID | Region | Latitude (degrees) | Longitude (degrees) | Description |
|---|---|---|---|---|
| D01 | CBD | 1.287693895 | 103.8514652 | Asian Civilisations Museum |
| D02 | East | 1.321819702 | 103.9144639 | Jalan Senyum (Road) |
| D03 | East | 1.342467129 | 103.9633338 | Singapore University of Technology and Design Staff Housing |
| D04 | East | 1.372735405 | 103.9496974 | White Sands Shopping Mall |
| D05 | Central | 1.305542629 | 103.8222091 | Napier Road |
| D06 | Central | 1.336814235 | 103.7931607 | The Grandstand Shopping Mall |
| D07 | Central | 1.344033838 | 103.8470656 | Bishan Harmony Park |
| D08 | North | 1.407594732 | 103.7576143 | Mandai Estate |
| D09 | North | 1.417440810 | 103.8332204 | Khatib MRT Station |
| D10 | North | 1.443074739 | 103.7904874 | Woodlands North Plaza |
| D11 | North | 1.448458900 | 103.8223306 | Intersection of Canberra Road & Old Nelson Road |
| D12 | North-east | 1.358085773 | 103.8887448 | Hougang Block 236 (Residential Building) |
| D13 | North-east | 1.399313682 | 103.8852278 | Sengkang Riverside Park |
| D14 | West | 1.282380986 | 103.6306377 | Tuas South Avenue 7 |
| D15 | West | 1.321123540 | 103.7405868 | Teban Neighborhood Park |
| D16 | West | 1.332750479 | 103.6394783 | Tuas West Road MRT Station |
| D17 | West | 1.336054064 | 103.6840244 | Singapore Discovery Centre (Museum) |
| D18 | West | 1.391549335 | 103.6987229 | Lim Chu Kang Road |

In addition, no in-situ verification has been carried out at the characteristic soundscapes identified yet, since participants answered from memory in the online questionnaire described in Section 3.3, and we performed the weighted *k*-means clustering purely based on the locations provided by the participants. It may not be practical to conduct soundwalks or record the soundscapes at those locations, and future in-situ studies using the characteristic soundscape locations may need to use the closest *feasible* location to the identified cluster centers instead of just the closest location described in Section 5.2.

## 6. Conclusions and Future Work

In conclusion, we conducted a questionnaire with 67 participants to obtain their opinions on soundscapes in Singapore that are characteristically "full of life and exciting", "chaotic and restless", "calm and tranquil", and "lifeless and boring". With the locations chosen by the participants, we implemented a weighted *k*-means clustering algorithm to identify a selection of characteristic soundscapes for each set of perceptual attributes, with the weights encoding the reliability (in the form of a probability of accurate identification) of the location. The locations identified to have characteristic soundscapes were evenly distributed around mainland Singapore, and were distinct for different sets of perceptual attributes.

Since the participants were local experts, the application of the questionnaire served as a bottom-up approach to site identification for soundscape studies, in contrast to standard top-down approaches to site identification adopted by soundscape researchers. Our proposed clustering method was carried out without external input from us as study researchers, and did not assume prior knowledge of Singapore, so the method effectively blinded us against bias in site selection and can also be applied to other countries or regions in a replicable manner.

Under our proposed method, each location identified to have a characteristic soundscape could also be interpreted as a "soundmark", defined by Schafer as "a community sound which is unique or possesses qualities which make it specially regarded or noticed by the people in that community" [39]. Hence, the locations identified in our study (or using our method) can also inform urban planners about important places whose soundscapes might be valuable to preserve and conserve when designing or redesigning a given venue or town [40,41], thereby allowing for better-guided sustainable development strategies for the venue or town. This is in line with United Nations Sustainable Development Goal 11.3 [42], which aims to "enhance inclusive and sustainable urbanization and capacity for participatory, integrated and sustainable human settlement planning and management".

Natural extensions for future work would thus involve the recording of the physical locations into an audio-visual database similar to the USotW database, with in-situ responses obtained via a standard protocol such as the SSID Protocol outlined by Mitchell and colleagues [43]. Longitudinal studies observing or recording the soundscapes over longer periods of time could also be conducted with the installation of wireless acoustic sensor networks (such as those described in [44–46]) with nodes installed at the identified locations.

Recording the characteristic soundscapes as audio-visual media furthers the goal of soundscape conservation and has the twofold effect allowing faithful reproduction in laboratory conditions, thereby increasing the ecological validity of results obtained using said recordings in the local context. The recordings can also serve as an important reference for urban planners and soundscape researchers when comparing or classifying other locations that have soundscapes that are similar in nature to, but are not part of, the representative set.

**Supplementary Materials:** NIL.

**Author Contributions:** Conceptualization, methodology and visualization, K. Ooi, B. Lam, and K.N. Watcharasupat; software and formal analysis, K. Ooi; validation, investigation, and data curation, K. Ooi and Z.T. Ong; writing—original draft preparation, K. Ooi; writing—review and editing, K. Ooi, B. Lam, J.Y. Hong, K.N. Watcharasupat, Z.T. Ong, and W.S. Gan; resources, supervision, project administration, and funding acquisition, J.Y. Hong and W.S. Gan. All authors have read and agreed to the published version of the manuscript.

**Funding:** This research/project is supported by the National Research Foundation, Singapore, and Ministry of National Development, Singapore under its Cities of Tomorrow R&D Programme (CoT Award: COT-V4-2020-1). Any opinions, findings and conclusions or recommendations expressed in this material are those of the author(s) and do not reflect the view of National Research Foundation, Singapore and Ministry of National Development, Singapore.

**Institutional Review Board Statement**: Prior ethical approval for conducting the study was obtained from the Institutional Review Board, Nanyang Technological University (Reference number IRB-2021-296).

**Informed Consent Statement**: Informed consent was obtained from all participants before they were included in the study. Participants were free to withdraw at any point in the study by exiting the browser window in which the online questionnaire was carried out.

**Data Availability Statement:** The anonymized raw data obtained from the participants, as well as code to replicate the clustering approach used to obtain the final characteristic soundscapes is available at https://github.com/ntudsp/singapore-soundscape-site-selection-survey. A Google Maps visualization of the raw data and final clusters, as shown in Figure 6, is available at https://www.google.com/maps/d/u/0/edit?mid=16fjoOwG-AnmwhTfc4MR11DrL_6iDL979.

**Acknowledgments:** The authors would like to thank Kelvin Lim (School of Electrical and Electronic Engineering, Nanyang Technological University) and Alroy Chiang (School of Physical and Mathematical Sciences, Nanyang Technological University) for testing out and providing insightful suggestions to streamline the online survey form.

**Conflicts of Interest:** The authors declare no conflict of interest. The funders had no role in the design of the study; in the collection, analyses, or interpretation of data; in the writing of the manuscript, or in the decision to publish the results.

## Abbreviations

The following abbreviations are used in this manuscript:

| | |
|---|---|
| API | Application Programming Interface |
| AUC | Area under curve |
| CBD | Central Business District |
| GPS | Global Positioning System |
| ISO | International Organization for Standardization |
| MRT | Mass rapid transit (train network) |
| NTU | Nanyang Technological University (Singapore) |
| SD | Standard deviation |
| STB | Singapore Tourism Board |
| URA | Urban Redevelopment Authority (Singapore) |
| USotW | Urban Soundscapes of the World (database) |
| WNSS | Weinstein Noise Sensitivity Scale |

**Appendix: Questionnaire**

The questionnaire consisted of 24 questions, each corresponding to one of the pairwise combinations of perceptual attributes in the 4 quadrants of the ISO 12913-2 circumplex model (full of life and exciting, chaotic and restless, calm and tranquil, lifeless and boring) and 6 major planning regions of Singapore (Central Area ("CBD Area"), Central Region, East Region, North Region, North-east Region, West Region). Each question consisted of the following parts:

A. Considering public open spaces (e.g. streets, squares, parks, etc.) in the <*Planning Region*> of Singapore, where do you experience the soundscape to be most <*Perceptual Attributes*>?
B. Coordinates of your chosen location.
C. Please explain and elaborate on your choice of location. For example, you can indicate the typical time and day of the week, etc. that you find the location to have a soundscape that is <*Perceptual Attributes*>.
D. How often do you visit your chosen location, or pass by it on foot?
E. How many times have you visited your chosen location, or passed by it on foot?
F. On average, how long do you spend at your chosen location, or pass by it on foot?

All questionnaires were administered via an online platform. For part A, participants were given a virtual map where they could drag a marker to their desired location to denote their response. Participants were advised not to choose a location that they have never visited or passed by on foot before, and markers could also be placed by entering names of locations in a search bar on the map. In addition, there was an internal check to ensure that all markers were dragged at least once, to ensure that markers were not accidentally left at their default locations, and participants were also informed of this fact. For reference, the default marker locations are as shown in Table A1 and are Mass Rapid Transit (MRT) stations (train stations) located roughly in the center of their corresponding regions.

**Table A1.** Default marker locations for virtual maps in online questionnaire

| Region | Latitude (degrees) | Longitude (degrees) | Description |
|---|---|---|---|
| Central Area ("CBD Area") | 1.2830173 | 103.8513365 | Raffles Place MRT Station |
| Central Region | 1.3199584 | 103.8259427 | Stevens MRT Station |
| East Region | 1.3532359 | 103.9452235 | Tampines MRT Station |
| North Region | 1.4273512 | 103.7931482 | Woodlands South MRT Station |
| North-east Region | 1.3829481 | 103.8933582 | Buangkok MRT Station |
| West Region | 1.3376415 | 103.6968990 | Pioneer MRT Station |

For part B, the coordinates of the participants' chosen location were automatically populated by the online questionnaire to match the marker location in part A. The answers to parts C and D were not used for this study, but may be used for future studies. A screenshot of a sample page of the online questionnaire is shown in Figure A1.

**Q1A)** Considering public open spaces (e.g. streets, squares, parks, etc.) in the CBD Area of Singapore, where do you experience the soundscape to be the most **full of life and exciting?**

Please *drag the marker* on the map to that location (zoom in/out if needed).
Please *do not* choose a location that you have never visited or passed by on foot before.
If the original location of the marker (Raffles Place MRT Station) *is* your chosen location, please drag the marker somewhere nearby first, then drag it back to its original location as our system may return an error if the marker has not been moved at all.
You may type the name of a location in the "Search Address" bar (embedded in the map) and press "Enter" to automatically move the marker to that location, but please ensure that the marker location is correct before submitting.

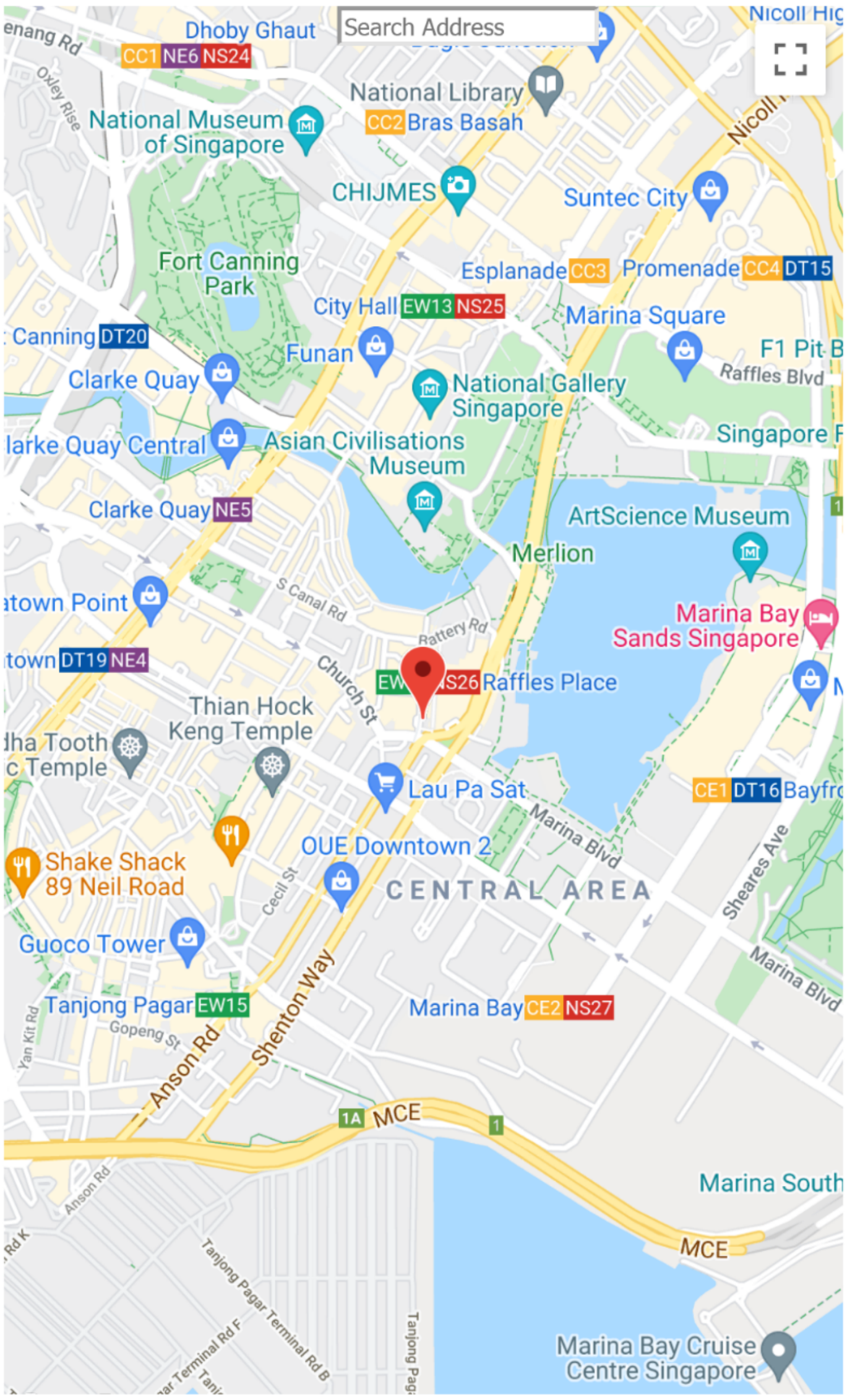

**Q1B)** Coordinates of your chosen location (please do not change these values; they will update when you drag the marker on the map):*

**Q1C)** Please explain and elaborate on your choice of location in Q1A). For example, you can indicate the typical time and day of the week, etc. that you find the location to have a soundscape that is **full of life and exciting**. You may also enter the location's name (if it has a name).*

**Q1D)** How often do you visit your chosen location in Q1A), or pass by it on foot?*

- Daily (i.e. about once a day)
- Between daily to weekly (i.e. about 2-6 times a week)
- Weekly (i.e. about once a week)
- Between weekly to monthly (i.e. about 2-3 times a month)
- Monthly (i.e. about once a month)
- Quarterly (i.e. about once every 3 months)
- Yearly (i.e. about once a year)
- Less often than once a year

**Q1E)** How many times have you visited your chosen location in Q1A), or passed by it on foot?*

If you chose anything *other* than "Less often than once a year" for Q1D), then you have visited your chosen location 10 or more times.

- 1 to 3 times
- 4 to 6 times
- 7 to 9 times
- 10 or more times

**Q1F)** On average, how long do you spend at your chosen location in Q1A) each time you visit it, or pass by it on foot?*

Please enter a time in minutes. E.g. if you spend roughly 5 minutes at your chosen location each time, enter "5". If you spend roughly 4 hours at your chosen location each time, enter "240".

**Figure A1.** Screenshot of sample page of online questionnaire administered to participants